\documentclass[10pt,twocolumn,letterpaper]{article}

\usepackage[pagenumbers,datasets]{wacv}%

\usepackage{multirow}
\usepackage{rotating}
\usepackage{graphicx}
\usepackage{longtable}
\usepackage{booktabs}
\usepackage{array}
\usepackage{graphicx}
\usepackage{placeins}
\usepackage{tabularx}
\usepackage{xspace}
\usepackage{amsmath,amsfonts,amssymb}
\DeclareMathAlphabet{\mathbbold}{U}{bbold}{m}{n}
\usepackage{threeparttable}
\newcolumntype{L}{>{\raggedright\arraybackslash}X}

\usepackage{listings}
\usepackage{tikz}
\usepackage{xcolor}
\usepackage{amssymb}
\definecolor{codegreen}{rgb}{0,0.6,0}
\definecolor{codegray}{rgb}{0.5,0.5,0.5}
\definecolor{codepurple}{rgb}{0.58,0,0.82}
\definecolor{backcolour}{rgb}{0.95,0.95,0.92}

\lstdefinestyle{mystyle}{
    backgroundcolor=\color{backcolour},
    commentstyle=\color{codegreen},
    keywordstyle=\color{magenta},
    numberstyle=\tiny\color{codegray},
    stringstyle=\color{codepurple},
    basicstyle=\ttfamily\footnotesize,
    xleftmargin=10pt,
    breakatwhitespace=false,
    breaklines=true,
    captionpos=b,
    keepspaces=true,
    numbers=left,
    numbersep=5pt,
    showspaces=false,
    showstringspaces=false,
    showtabs=false,
    tabsize=2
}

\definecolor{wacvblue}{rgb}{0.21,0.49,0.74}
\usepackage[pagebackref,breaklinks,colorlinks,allcolors=wacvblue]{hyperref}

\def\wacvPaperID{2409}%
\def\confName{WACV}
\def\confYear{2027}

\title{CounterFactual Retrieval for Evidence-Aware Individual Animal Identification}

\author{
Ankit K. Upadhyay$^{1}$,
Aleksandr Algasov$^{2,3}$,
Ekaterina Nepovinnykh$^{2}$,
Fedor Zolotarev$^{4}$, \\
Tuomas Eerola$^{2}$,
Heikki K\"{a}lvi\"{a}inen$^{2,3}$,
Charles V. Stewart$^{1}$,
Lukas Picek$^{5,6}$ \vspace{1mm} \\
$^{1}$Rensselaer Polytechnic Institute
$^{2}$LUT University,
$^{3}$Brno University of Technology, \\
$^{4}$University of Helsinki,
$^{5}$MIT,
$^{6}$University of West Bohemia
}

\begin{document}

\maketitle

\begin{abstract}
Automated identification of individual animals is moving from small curated studies toward shared benchmarks and reusable foundation models. This progress is useful, but current evaluation leaves the evidence behind a match unresolved, i.e., high retrieval accuracy does not show whether a model used animal features or identity-correlated shortcuts.
We address this with Counterfactual Foreground Retrieval (CFR), an evidence-aware evaluation protocol.
CFR evaluates the same query-gallery task under different views: full image, foreground-only, and shortcut-only.
The full-image measures standard retrieval performance;
the foreground-only measures foreground-supported retrieval;
and shortcut-only view measure how much identity signal is recoverable from background, or silhouette.
This turns a retrieval score into an evidence profile: how well the model performs on the original images, how much of that performance remains when only the animal is visible, and how much identity signal remains recoverable when the animal is removed.
We release \href{https://huggingface.co/datasets/imageomics/cfr-wildlife-masks}{around 0.5M masks for 34 datasets} and \href{https://github.com/Imageomics/cfr}{code for mask generation and CFR evaluation}.\footnote{\url{https://github.com/Imageomics/cfr}}
\end{abstract}

\vspace{-2mm}
\section{Introduction}
\label{sec:intro}

Identifying individual animals across time and space is central to many ecological studies, e.g., population estimation, survival, and movement~\cite{lukacs2005review,schofield2020long}. Automating this task is increasingly important as manual photo-identification is slow, unreliable, labor-intensive, and hard to scale~\cite{vidal2021perspectives}. As a result, the field is moving from proprietary design and deployment towards open benchmarks and reusable foundation models~\cite{vcermak2024wildlifedatasets,otarashvili2024multispecies}.
This shift has made methods easier to compare and potentially deploy, but it has also exposed a weakness in their evaluation. A correct retrieval does not necessarily mean that the model used the biometric the dataset was built around, such as a coat pattern, an arrangement of facial scales, or a fin contour. It may also be driven by background, capture context, near-duplicates, training-set overlap, or by body outline in species whose identity comes from a surface pattern (see Tab. \ref{tab:shortcut-evidence}).
\begin{figure}
\vspace{0mm}
    \centering
        \includegraphics[width=\linewidth]{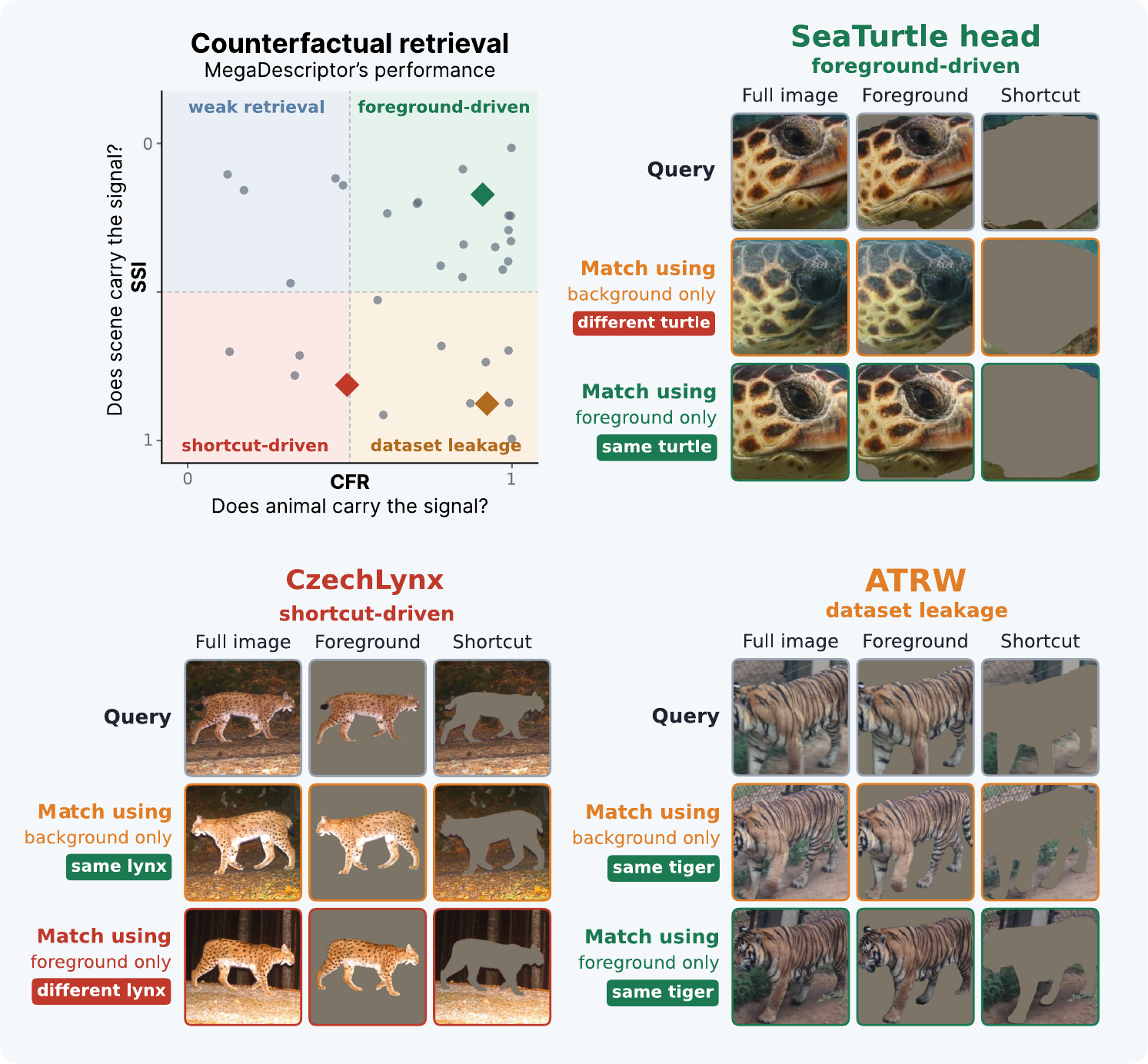}
    \caption{\textbf{Strong retrieval scores depend on different evidence.} Standard retrieval reports whether the correct individual is retrieved, but not which visual evidence supports it. We therefore introduce the Shortcut Signal Indicator (SSI), which measures how much of the full-image retrieval signal can be recovered from \textit{shortcut-only} inputs. Among otherwise strong results, SSI ranges from almost zero to almost one, showing why retrieval performance should be evaluated along with the evidence behind it.}
    \label{fig:placeholder}
\end{figure}

This ambiguity matters because the settings in which the methods for automated identification are used are usually far from the benchmark. Field images vary across cameras, observers, locations, seasons, and capture conditions. If an evaluation rewards context that correlates with identity, its accuracy can overstate the model's reliability once deployed. \textbf{The core evaluation question should therefore not only be whether a model retrieves the correct individual, but what evidence makes it possible.}

\newpage
We address this with \emph{Counterfactual Foreground Retrieval} (CFR), a leakage-aware evaluation protocol.
CFR evaluates the same query-gallery retrieval task on matched full-image, foreground-only, and shortcut-only. These views separate standard retrieval performance from animal-visible evidence and shortcut evidence. CFR summarizes how much of the full-image retrieval score remains supported by the foreground, while the shortcut signal indicator (SSI) measures how much of the full-image signal is recoverable from the background or from the animal's body outline, which we isolate with a silhouette-only view. Outline is excluded from the shortcut signal for species whose identity is carried by shape, such as whale flukes and dorsal fins. CFR does not replace full-image accuracy or treat foreground-only retrieval as leakage-free; together, CFR and SSI report the evidence profile behind the score.

Alongside the CFR protocol, we update the WildlifeDatasets library~\cite{vcermak2024wildlifedatasets} into a leakage-auditable benchmark suite. We add the missing segmentation and evaluation layer on top of existing datasets: foreground masks for roughly 0.5M images from 34 datasets, mask-generation code for future use, and CFR-related evaluation code. All this allows testing whether retrieval is supported by animal-visible evidence, background context, or silhouette cues, whereas previously, only full-image retrieval was reported.

Finally, we use CFR to audit selected models on a wide variety of datasets. The results show that full-image Retrieval@$k$ can hide very different evidence profiles. Some model–dataset pairs are strongly foreground-supported, with high CFR and low SSI. Others are overdetermined, i.e., the animal foreground supports retrieval, but background or silhouette cues are also predictive. In the most confounded cases, shortcut-only inputs recover a large fraction of the full-image signal. These results show why full-image accuracy is not enough. It reports only whether retrieval succeeds, but not whether the success comes from the animal or from the dataset shortcuts.

\medskip
\noindent\textbf{Contributions.}
Our work shifts the existing evaluation of individual animal identification towards a more reliable one:

\begin{itemize}
\item \textbf{We show that standard scores can be incomplete.}
We provide evidence across multiple datasets and models, that full-image retrieval hides shortcut reliance by retaining identity signals from background and/or silhouette.
\vspace{1mm}
\item \textbf{We introduce Counterfactual Foreground Retrieval.}
CFR evaluates the same retrieval task under matched full-image, foreground- and shortcut-only views, separating operational performance from animal-visible evidence and shortcut signal.

\vspace{1mm}
\item \textbf{We make existing benchmarks auditable.}
We add accurate masks to around 0.5M images across 34 publicly available datasets and code to generate them, enabling CFR where it was previously not possible.

\end{itemize}

\begin{table}[!h]
\begin{threeparttable}
\vspace{1mm}
\centering
\scriptsize
\setlength{\tabcolsep}{3.8pt}
\begin{tabular}{llcccc}
\toprule
\textbf{Dataset} & \textbf{Model} & \textbf{Random} & \textbf{Full image} & \textbf{Foreg.} & \textbf{Shortcut} \\
\midrule
\multirow{2}{*}{WhaleSharkID~\cite{holmberg2009estimating}} & DINOv3 & 0.5 & 35.3 & 30.3 & 19.1 \\
 & MegaD$^C$ & 0.5 & 30.5 & 24.8 & 16.3 \\
\midrule
\multirow{2}{*}{LeopardID2022~\cite{leopardid2022}} & DINOv3 & 1.1 & 57.4 & 52.2 & 46.3 \\
 & MegaD$^C$ & 1.1 & 51.6 & 44.0 & 40.5 \\
\midrule
\multirow{2}{*}{Saimaa seal~\cite{saimaa_wildbook_2026}} & DINOv3 & 0.7 & 46.6 & 24.4 & 39.1 \\
 & MegaD$^C$ & 0.7 & 36.7 & 28.3 & 28.9 \\
\midrule
\multirow{2}{*}{SeaTurtleID2022~\cite{adam2024seaturtleid2022}} & DINOv3 & 0.5 & 59.0 & 57.3 & 18.8 \\
 & MegaD$^C$ & 0.5 & 60.6 & 58.6 & 13.9 \\
\midrule
\multirow{2}{*}{StripeSpotter~\cite{lahiri2011biometric}} & DINOv3 & 3.0 & 82.1 & 81.6 & 3.2 \\
 & MegaD$^C$ & 3.0 & 96.6 & 96.5 & 3.3 \\
\midrule
\multirow{2}{*}{HyenaID2022~\cite{hyenaid2022}} & DINOv3 & 0.7 & 63.7 & 60.6 & 44.1 \\
 & MegaD$^C$ & 0.7 & 62.8 & 52.4 & 38.8 \\
\midrule
\multirow{2}{*}{BrownBearHeads~\cite{rosenberg2026individual}} & DINOv3 & 1.6 & 99.0 & 98.7 & 72.4 \\
 & MegaD$^C$ & 1.6 & 93.6 & 91.0 & 66.6 \\
\midrule
\multirow{2}{*}{CzechLynx~\cite{picek2026czechlynx}} & DINOv3 & 1.9 & 68.1 & 55.7 & 56.6 \\
 & MegaD$^C$ & 1.9 & 60.4 & 48.6 & 45.7 \\
\midrule
\multirow{2}{*}{NyalaData~\cite{dlamini2020automated}} & DINOv3 & 0.7 & 14.8 & 15.8 & 5.9 \\
 & MegaD$^C$ & 0.7 & 15.2 & 13.9 & 6.4 \\
\bottomrule
\end{tabular}
\end{threeparttable}
\caption{\textbf{Standard evaluation hides shortcut signal.} Even in the zero-shot setting, the shortcut-only view can retrieve correct identities above random chance. This indicates that full-image Retrieval@$k$ can reflect not only animal-visible evidence, but also identity-correlated background or silhouette cues. $C$--not trained on any datasets in this table.}
\label{tab:shortcut-evidence}
\vspace{-2mm}
\end{table}

\section{Related Work}
\label{sec:related}

Animal re-identification has moved from matching hand-designed visual patterns to learning reusable embeddings. Early systems such as StripeCodes~\cite{lahiri2011biometric}, HotSpotter~\cite{crall2013hotspotter}, MantaMatcher~\cite{town2013m}, the modified Groth algorithm~\cite{arzoumanian2005astronomical}, and ALFRE-ID~\cite{nepovinnykh2024species} relied on local features, geometric verification, and database search over markings such as stripes, spots, scars, fins, and contours. Recent work scales this idea with larger benchmarks (e.g., WildlifeReID-10k~\cite{adam2025wildlifereid}) and multi-species models, including MegaDescriptor~\cite{vcermak2024wildlifedatasets}, MiewID~\cite{otarashvili2024multispecies}. This progress has made animal re-ID easier to benchmark and reuse, but the evaluation signal remains narrow: rank-$k$ accuracy and mAP say whether retrieval worked, not what visual evidence made it work.

\medskip
\noindent\textbf{Leakage in benchmark design.}
The limits of rank-$k$ accuracy are especially clear when benchmark construction leaks information across evaluation examples. Prior work has addressed this mainly through better split design, including similarity-aware protocols that reduce overlap from repeated, near-duplicate, or visually similar captures~\cite{adam2024seaturtleid2022,adam2025wildlifereid}.

\medskip
\noindent\textbf{Foreground, background, and shortcut cues.}
The same issue also appears at the pixel level and models can use visual correlations that are predictive in a benchmark but outside the intended identity signal~\cite{geirhos2020shortcut}. Background is a well-known example. It can support prediction on its own, and changing it can change model decisions even when the foreground object is fixed~\cite{xiao2021noise}. In animal re-ID, recent work has addressed this by removing backgrounds, modeling foreground and background separately, or diagnosing whether retrieval relies on coat pattern, background, or laterality cues~\cite{yu2024elephant,picek2024fgbg,ruedatoicen2026jaguar}. Related foreground/background studies in wildlife behavior recognition show the same pattern outside individual identification~\cite{brookes2025panaf}.

\medskip
\noindent\textbf{Segmentation masks for benchmark auditing.}
Separating animal-visible evidence from surrounding context in an existing benchmark requires a foreground mask for every evaluated image. Most animal re-ID benchmarks, however, were built for retrieval evaluation and provide images, identity labels, and evaluation protocols rather than segmentation annotations. A few datasets include masks or part annotations, such as SeaTurtleID2022~\cite{adam2024seaturtleid2022}, SealID~\cite{nepovinnykh2022sealid}, and CzechLynx~\cite{picek2026czechlynx}, but this is not yet standard. Promptable segmentation with SAM has made mask generation more accessible~\cite{kirillov2023segment}, with SAM3 extending this direction to concept-prompted segmentation from text or visual exemplars~\cite{sam3}.
\section{Counterfactual Foreground Retrieval}
\label{sec:cfr}

Standard evaluation with full-image Retrieval@$k$ measures whether the correct individual is retrieved, but not what visual evidence made the retrieval possible. With that in mind, we introduce \textbf{C}ounterfactual \textbf{F}oreground \textbf{R}etrieval to make such evidence more explicit. CFR keeps the task fixed and changes only the visual input. For each image, we construct matched counterfactual views (full image, foreground-only, and shortcut-only) under the same identity labels, query-gallery structure, model, and scoring protocol.

\subsection{Evaluation Setup}
\label{sec:cfr-task}

Let $\mathcal{D}=\{(x_i,y_i,m_i)\}_{i=1}^{N}$ be an evaluation dataset, where $x_i \in \mathbb{R}^{H \times W \times 3}$ is an image crop, $y_i$ is the individual identity label, and $m_i \in \{0,1\}^{H \times W}$ is a binary foreground mask for the animal. Let $\mathcal{Q}$ be the set of valid query images. For each query $i \in \mathcal{Q}$, let $\mathcal{G}_i$ be the valid gallery under the benchmark protocol. The gallery excludes the query annotation itself and follows the same validity rules as the standard evaluation. The positive set for query $i$ is
\begin{equation}
\mathcal{P}_i = \{j \in \mathcal{G}_i : y_j = y_i\}.
\end{equation}
We evaluate only queries with $|\mathcal{G}_i|>0$ and $|\mathcal{P}_i|>0$. CFR keeps $\mathcal{Q}$, $\mathcal{G}_i$, $\mathcal{P}_i$, identity labels, and validity rules fixed across all counterfactual inputs.

\subsection{Counterfactual Inputs}
\label{sec:cfr-inputs}

The counterfactual inputs isolate different sources of retrieval evidence while keeping the same images, labels, and query-gallery protocol. Each original image is transformed into a foreground-only view and one or more shortcut-only views (see Fig.~\ref{fig:cfr_examples}). The foreground-only view keeps animal-visible pixels and removes the surrounding image. Shortcut-only views remove animal texture while retaining non-foreground cues such as scene context or silhouette.

Let $\tilde{m}i \in {0,1}^{H \times W \times 3}$ be $m_i$ broadcast across color channels, and let $c \in \mathbb{R}^{H \times W \times 3}$ be a constant neutral fill image. The full-image input is
\begin{equation}
T_{\mathrm{full}}(x_i,m_i)=x_i.
\end{equation}
The foreground-only input is
\begin{equation}
T_{\mathrm{fg}}(x_i,m_i) = \tilde{m}_i \odot x_i + (1-\tilde{m}i)\odot c.
\end{equation}
The texture-removed shortcut input is
\begin{equation}
T_{\mathrm{sc}}(x_i,m_i) = (1-\tilde{m}_i)\odot x_i + \tilde{m}_i \odot c.
\end{equation}
The silhouette-only view is
\begin{equation}
T_{\mathrm{sil}}(x_i,m_i) = \tilde{m}_i \odot s + (1-\tilde{m}_i)\odot c,
\end{equation}
where $s \in \mathbb{R}^{H \times W \times 3}$ is a constant silhouette fill image. We construct this view for every dataset; Sec.~\ref{sec:cfr-signal} fixes whether it counts as shortcut evidence.\footnote{For species commonly identified by body outline, such as elephant ears or whale flukes, outline is part of the intended biometric and the silhouette-only view is omitted from the shortcut signal.}

\subsection{Retrieval Under Counterfactual Views}
\label{sec:cfr-base-score}

Each counterfactual view is evaluated with the same retrieval protocol. Let $T$ denote one of the CFR input views defined in Sec.~\ref{sec:cfr-inputs}: full image, foreground-only, texture-removed shortcut, or, when valid, silhouette-only. Let $f$ be the model being evaluated, and let $a(\cdot,\cdot)$ be its retrieval scoring function, with larger values indicating more similar images. For query $i$ and gallery image $j \in \mathcal{G}_i$, the view-specific retrieval score is
\begin{equation}
r_T(i,j) =
a\left(
f(T(x_i,m_i)),
f(T(x_j,m_j))
\right),
\end{equation}
where $r_T(i,j)$ is used to rank gallery image $j$ for query $i$ under view $T$.
After ranking gallery images in decreasing order of $r_T(i,j)$, let $\operatorname{TopK}T(i)$ denote the top-$k$ ranked gallery set. The top-$k$ retrieval rate under view $T$ is
\begin{equation}
\small
R_T@k =
\frac{1}{|\mathcal{Q}|}
\sum{i \in \mathcal{Q}}
\mathbf{1}
\left[
\exists j \in \operatorname{TopK}T(i)
\ \mathrm{s.t.}
y_j = y_i
\right].
\end{equation}

\textit{In the following, $R_{\mathrm{full}}@k$, $R_{\mathrm{fg}}@k$, $R_{\mathrm{sc}}@k$, and $R_{\mathrm{sil}}@k$ denote retrieval rates for the full-image, foreground-only, shortcut-only, and silhouette-only views, respectively.
}

\subsection{Random Retrieval Rate}
\label{sec:cfr-random}

Raw retrieval rates are not directly comparable across datasets because gallery size and the number of positives per query can vary. We therefore compute a random baseline using the same query-gallery structure as the evaluated model. For each query $i$, let $K_i=\min(k,|\mathcal{G}i|)$. A random ranking succeeds if at least one of the $|\mathcal{P}i|$ positives appears in the top $K_i$ positions.
Averaging this probability over valid queries gives
\begin{equation}
R{\mathrm{rand}}@k =
\frac{1}{|\mathcal{Q}|}
\sum{i\in\mathcal{Q}}
\left[
1 -
\frac{
\binom{|\mathcal{G}_i|-|\mathcal{P}_i|}{K_i}
}{
\binom{|\mathcal{G}_i|}{K_i}
}
\right],
\end{equation}
where the numerator is set to zero when $|\mathcal{G}i|-|\mathcal{P}i| < K_i$. For $k=1$, this reduces to
\begin{equation}
R{\mathrm{rand}}@1 =
\frac{1}{|\mathcal{Q}|}
\sum{i\in\mathcal{Q}}
\frac{|\mathcal{P}_i|}{|\mathcal{G}_i|}.
\end{equation}

\subsection{Foreground and Shortcut Signal}
\label{sec:cfr-signal}

Raw retrieval rates depend on the query-gallery structure, so we express each view-specific score relative to the random baseline from Sec.~\ref{sec:cfr-random}:
\begin{equation}
S_T@k = \max\left(0, R_T@k - R_{\mathrm{rand}}@k\right).
\end{equation}
This removes expected random retrieval and clips below-random values to zero. The full-image and foreground signals are
$S_{\mathrm{full}}@k = S_{T_{\mathrm{full}}}@k, S_{\mathrm{fg}}@k = S_{T_{\mathrm{fg}}}@k$.
The shortcut signal is computed from shortcut-only inputs:
\begin{equation}
S_{\mathrm{shortcut}}@k =
\begin{cases}
S_{T_{\mathrm{sc}}}@k, \\
\max\left(S_{T_{\mathrm{sc}}}@k, S_{T_{\mathrm{sil}}}@k\right).
\end{cases}
\end{equation}

The first case is used when silhouette is not considered a shortcut. The second is used when silhouette is treated as a shortcut. This choice is fixed per dataset based on the intended biometric rather than the model. Datasets identified by surface patterns use the second ($\max$) case, whereas datasets identified by outline (e.g., whale flukes) use the first case and report silhouette separately. Appendix~\ref{app:dataset-characteristics} lists the silhouette role for each dataset.

\subsection{CFR Score}
\label{sec:cfr-score}

We define CFR around a simple requirement: a retrieval score should remain high when the model is forced to rely on the animal foreground. Full-image retrieval gives the standard benchmark score, but it does not show whether the match was supported by the animal, by shortcut cues, or by both. CFR keeps the full-image score in the metric, but requires it to be matched by foreground-only retrieval.

Before combining the two views, we normalize their above-random signals by the available headroom:
\begin{equation}
\small
A_{\mathrm{full}}@k =
\frac{S_{\mathrm{full}}@k}{1 - R_{\mathrm{rand}}@k},
\qquad
A_{\mathrm{fg}}@k =
\frac{S_{\mathrm{fg}}@k}{1 - R_{\mathrm{rand}}@k}.
\end{equation}
The CFR score is then defined for $R_{\mathrm{rand}}@k < 1$ as
\begin{equation}
\mathrm{CFR}@k =
\min\left(A_{\mathrm{full}}@k, A_{\mathrm{fg}}@k\right).
\label{eq:cfr}
\end{equation}

The minimum makes the score conservative. A method receives high CFR only when it retrieves well under the original full-image input and when retrieval remains strong under the foreground-only input. If full-image retrieval is weak, CFR is low. If full-image retrieval is high but foreground-only retrieval drops, CFR is also low. Thus, CFR measures foreground-supported retrieval performance rather than foreground-only performance alone.

CFR does not subtract or penalize shortcut signal. High shortcut-only retrieval can mean that the benchmark contains identity-correlated context, even when the model also retrieves correctly from animal-visible evidence. We therefore report shortcut recoverability separately with the shortcut signal indicator:
\begin{equation}
\mathrm{SSI}@k =
\min\left(1,\frac{S_{\mathrm{shortcut}}@k}{S_{\mathrm{full}}@k}\right),
\label{eq:ssi}
\end{equation}
defined for $S_{\mathrm{full}}@k > 0$. SSI measures how much of the full-image retrieval signal is recoverable from shortcut-only inputs. Values near zero indicate little shortcut recoverability, while values near one indicate that shortcut-only inputs recover an identity signal comparable to the full image.
SSI is therefore just an evidence-confounding indicator, not a direct measure of method failure.

\subsection{Interpretation}
\label{sec:cfr-interpretation}

The proposed evaluation protocol should be read in the same way as standard Retrieval@$k$, with one important difference. Retrieval@$k$ reports performance under the original benchmark, whereas CFR and SSI repeat the same evaluation after controlled interventions that selectively remove different sources of visual evidence.
The resulting evidence profile distinguishes changes in retrieval performance caused by removing the animal from those caused by removing the surrounding visual context.
\textbf{Neither CFR nor SSI should be interpreted in isolation.} A decrease in CFR is informative only together with SSI. For example, a lower CFR accompanied by low SSI indicates that removing the foreground reduces retrieval while shortcut information remains uninformative. In contrast, a similar decrease together with high SSI indicates that shortcut-only inputs preserve much of the original retrieval signal. Likewise, high SSI should not be interpreted as model failure. It simply indicates that the evaluated benchmark contains shortcut cues that are predictive under the proposed intervention.

Beyond that, the provided evidence profile can help prioritize future improvements. For example, evidence-confounded and shortcut-driven profiles indicate the need for better train/test splits, more diverse data collection, explicit foreground-aware training, or benchmark curation that reduces identity-correlated shortcut cues. We evaluate some of the ideas below.

\begin{figure*}[t]
\centering
\includegraphics[width=\textwidth]{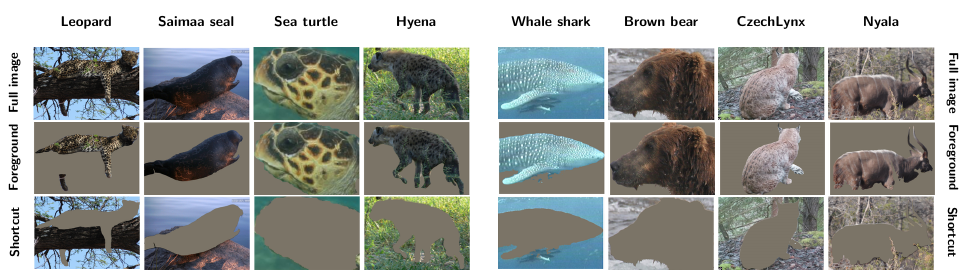}
\caption{\textbf{CounterFactual views.}
For each image, the provided mask allows us to generate matched full-image,
foreground-only, and shortcut-only inputs. Such views alter the available
visual evidence and enable evidence-aware evaluation.
}
\label{fig:cfr_examples}
\end{figure*}

\section{Benchmarking Suite}
\label{sec:benchmark-suite}

To enable CFR evaluation at scale, we provide segmentation masks for 34 datasets (roughly 0.5M annotations), release the code used to generate them, and substantially extend WildlifeDatasets~\cite{vcermak2024wildlifedatasets} so that masks can be loaded in a standardized way for all datasets along the original image, identity label, and annotation metadata.

\subsection{Segmentation Masks}
\label{sec:annotation-aligned-masks}

To enable high-quality mask generation at scale (see example masks in Fig. \ref{fig:cfr_examples}), we design a pipeline that combines prompt-based segmentation, conservative candidate selection, and manual quality assurance. Here we provide a short description; more details are available in Appendix~\ref{app:mask-processing}.

\medskip
\noindent\textbf{Mask generation.}
We generate foreground candidates using SAM3~\cite{kirillov2023segment,sam3}. For each dataset, we use a simple dataset-level prompt, such as the species name, a common group name, or the annotated body part for part-based datasets. This keeps mask generation scalable while allowing datasets focused on heads, flanks, or other body parts to use prompts that match the relevant visual evidence.

\medskip
\noindent\textbf{Mask acceptance.} We select one individual foreground mask for each image whenever a reliable candidate is available. If an annotation has a source bounding box, the selected mask must correspond to the annotated region rather than another animal or background object in the image. If the annotation is already a cropped image or full-image record, use the highest-scoring SAM3 candidate unless dataset-specific checks reject it. We retain rejected candidates and their metadata for audit and debugging. As masks and counterfactual views are constructed per annotation, we handle images with multiple animals or different annotated body parts by following the original annotations.%

\medskip
\noindent\textbf{Quality assurance.}
To verify that the selected masks support reliable foreground and background interventions, we compare them against 400 manually drawn reference masks (50 per random dataset). The agreement is high: median IoU is 98.2\%, mean IoU is 96.1\%, and 95.1\% of samples exceed 90\% IoU. Rejected low-confidence candidates provide an additional sanity check: many come from difficult images with occlusion, device artifacts, multiple animals, missing foreground, or ambiguous background.

\medskip
\noindent\textbf{Coverage.} Across the 34 datasets, \href{https://huggingface.co/datasets/imageomics/cfr-wildlife-masks}{the final release} contains
accepted masks for $457{,}238$ of $484{,}674$ eligible annotations
($94.3\%$ coverage). Since the pipeline prioritizes mask quality over complete
coverage, annotations without an accepted mask are excluded from the main CFR
evaluation. For dataset-level coverage statistics, see
Appendix~\ref{app:mask-coverage}.

\medskip
\noindent\textbf{Resource Format and Loading.}
Accepted masks are released as compact, run-length-encoded, and in one JSON file per dataset. Counterfactual views are generated on demand from each image--mask pair, ensuring that all views are derived from the same original annotation.

\section{Evidence Profiles Revealed by CFR}
\label{sec:results}

We use the CFR-ready benchmarking suite to demonstrate what standard full-image retrieval scores hide. The goal is not to rank MiewID and MegaDescriptor by unseen-dataset generalization. Instead, the goal is diagnostic. When a model retrieves the correct individual, does the retrieval remain supported by animal signature, or just background?

\medskip
\noindent\textbf{Standard evaluation hides different evidence profiles.}
A central finding is that similar full-image Retrieval@1 values correspond to different sources of evidence. Some datasets have both high full-image retrieval and high CFR, meaning that the animal foreground is sufficient for identification (see Tab.~\ref{tab:all-cfr-miewid-megad}). This is the cleanest case for interpreting standard retrieval performance as animal-visible identity matching.
However, high full-image retrieval can also occur when shortcut-only views are predictive. OpenCows2020, AAUZebraFish, and ATRW have strong full-image performance and high CFR, but also high SSI. These datasets are therefore overdetermined: the animal foreground supports retrieval, but identity-correlated shortcut evidence as well. Standard Retrieval@1 alone would not distinguish that.

\begin{table}[!h]
\centering
\scriptsize
\setlength{\tabcolsep}{2.5pt}
\begin{tabular}[t]{@{}llrrrrrr@{}}
\toprule
\textbf{Dataset} & \textbf{Model} & \textbf{Random} & \textbf{Full} & \textbf{Frg.} & \textbf{SC} & \textbf{SSI} & \textbf{CFR} \\
\midrule
\multicolumn{8}{@{}l}{\textit{\textbf{High shortcut recoverability}}} \\
\midrule
\multirow{2}{*}{OpenCows2020~\cite{andrew2021visual}} & MiewID & 3.7 & 99.1 & 98.4 & 73.3 & 73.0 & 98.4 \\
 & MegaDes. & 3.7 & 100.0 & 99.0 & 70.9 & 69.9 & 99.0 \\
\midrule
\multirow{2}{*}{AAUZebraFish~\cite{bruslund2020re}} & MiewID & 16.7 & 98.3 & 97.6 & 90.2 & 90.1 & 97.1 \\
 & MegaDes. & 16.7 & 100.0 & 99.3 & 89.6 & 87.5 & 99.1 \\
\midrule
\multirow{2}{*}{ATRW~\cite{li2019atrw}} & MiewID & 1.2 & 94.1 & 91.0 & 86.0 & 91.4 & 90.9 \\
 & MegaDes. & 1.2 & 95.9 & 92.4 & 84.4 & 87.8 & 92.3 \\
\midrule
\multirow{2}{*}{FriesianCattle2017~\cite{andrew2017visual}} & MiewID & 2.2 & 90.6 & 84.5 & 76.2 & 83.7 & 84.1 \\
 & MegaDes. & 2.2 & 99.2 & 87.5 & 87.4 & 87.7 & 87.2 \\
\midrule
\multirow{2}{*}{CzechLynx~\cite{picek2026czechlynx}} & MiewID & 1.9 & 75.6 & 68.0 & 52.8 & 69.0 & 67.4 \\
 & MegaDes. & 1.9 & 61.3 & 50.2 & 50.3 & 81.5 & 49.2 \\
\midrule
\multirow{2}{*}{IPanda50~\cite{wang2021giant}} & MiewID & 2.5 & 56.1 & 46.3 & 52.3 & 92.9 & 44.9 \\
 & MegaDes. & 2.5 & 97.3 & 78.8 & 67.3 & 68.3 & 78.3 \\
\midrule
\multirow{2}{*}{Saimaa seal~\cite{nepovinnykh2022sealid}} & MiewID & 0.7 & 43.9 & 33.0 & 31.3 & 70.9 & 32.5 \\
 & MegaDes. & 0.7 & 41.5 & 33.5 & 32.6 & 78.3 & 33.0 \\
\midrule
\multirow{2}{*}{NDD20~\cite{trotter2020ndd20}} & MiewID & 1.6 & 45.8 & 29.9 & 35.4 & 76.4 & 28.7 \\
 & MegaDes. & 1.6 & 85.2 & 35.6 & 61.4 & 71.5 & 34.5 \\
\midrule
\multicolumn{8}{@{}l}{\textit{\textbf{Mixed evidence}}} \\
\midrule
\multirow{2}{*}{Cows2021~\cite{gao2021towards}} & MiewID & 0.7 & 98.7 & 98.3 & 27.3 & 27.1 & 98.3 \\
 & MegaDes. & 0.7 & 99.8 & 99.7 & 24.7 & 24.2 & 99.7 \\
\midrule
\multirow{2}{*}{Giraffes~\cite{miele2021revisiting}} & MiewID & 0.6 & 98.3 & 97.4 & 39.9 & 40.3 & 97.4 \\
 & MegaDes. & 0.6 & 98.4 & 97.2 & 42.1 & 42.4 & 97.2 \\
\midrule
\multirow{2}{*}{BrownBearHeads~\cite{rosenberg2026individual}} & MiewID & 1.6 & 96.5 & 96.0 & 62.6 & 64.2 & 95.9 \\
 & MegaDes. & 1.6 & 92.1 & 92.4 & 68.4 & 73.8 & 92.0 \\
\midrule
\multirow{2}{*}{HumpbackWhaleID~\cite{kaggle2019humpback}} & MiewID & 0.1 & 88.8 & 83.6 & 72.1 & 81.2 & 83.6 \\
 & MegaDes. & 0.1 & 88.2 & 84.8 & 28.4 & 32.2 & 84.8 \\
\midrule
\multirow{2}{*}{HappyWhale~\cite{kaggle2022happywhale}} & MiewID & 0.1 & 45.2 & 35.3 & 32.4 & 71.7 & 35.3 \\
 & MegaDes. & 0.1 & 63.1 & 31.8 & 29.7 & 47.0 & 31.7 \\
\midrule
\multirow{2}{*}{LeopardID2022~\cite{leopardid2022}} & MiewID & 1.1 & 92.5 & 89.1 & 38.9 & 41.3 & 89.0 \\
 & MegaDes. & 1.1 & 78.4 & 59.1 & 41.9 & 52.8 & 58.6 \\
\midrule
\multirow{2}{*}{HyenaID2022~\cite{hyenaid2022}} & MiewID & 0.7 & 87.4 & 83.8 & 37.4 & 42.4 & 83.7 \\
 & MegaDes. & 0.7 & 92.5 & 78.2 & 38.5 & 41.1 & 78.0 \\
\midrule
\multirow{2}{*}{CTai~\cite{freytag2016chimpanzee}} & MiewID & 2.7 & 94.0 & 89.2 & 32.5 & 32.6 & 88.9 \\
 & MegaDes. & 2.7 & 98.0 & 95.0 & 35.9 & 34.8 & 94.9 \\
\midrule

\multicolumn{8}{@{}l}{\textit{\textbf{Low shortcut recoverability}}} \\
\midrule
\multirow{2}{*}{MacaqueFaces~\cite{witham2018automated}} & MiewID & 2.9 & 98.9 & 98.2 & 15.3 & 12.8 & 98.2 \\
 & MegaDes. & 2.9 & 99.4 & 99.0 & 26.2 & 24.1 & 99.0 \\
\midrule
\multirow{2}{*}{WhaleSharkID~\cite{holmberg2009estimating}} & MiewID & 0.5 & 88.7 & 82.0 & 14.3 & 15.6 & 81.9 \\
 & MegaDes. & 0.5 & 83.4 & 70.9 & 17.2 & 20.1 & 70.7 \\
\midrule
\multirow{2}{*}{BelugaID~\cite{belugaid2022}} & MiewID & 0.4 & 77.4 & 64.0 & 10.3 & 12.8 & 63.9 \\
 & MegaDes. & 0.4 & 84.1 & 71.2 & 16.8 & 19.6 & 71.1 \\
\midrule
\multirow{2}{*}{Reunion turtles~\cite{adam2025exploiting}} & MiewID & 0.9 & 92.0 & 91.4 & 8.0 & 7.8 & 91.3 \\
 & MegaDes. & 0.9 & 61.9 & 64.0 & 15.2 & 23.4 & 61.6 \\
\midrule
\multirow{2}{*}{SeaTurtleID2022~\cite{adam2024seaturtleid2022}} & MiewID & 0.5 & 96.9 & 95.8 & 11.4 & 11.4 & 95.8 \\
 & MegaDes. & 0.5 & 92.4 & 91.0 & 16.1 & 17.0 & 91.0 \\
\midrule
\multirow{2}{*}{Amvrakikos turtles~\cite{adam2025exploiting}} & MiewID & 1.5 & 65.6 & 70.8 & 5.1 & 5.6 & 65.1 \\
 & MegaDes. & 1.5 & 49.7 & 48.7 & 8.2 & 13.9 & 47.9 \\
\midrule
\multirow{2}{*}{NyalaData~\cite{dlamini2020automated}} & MiewID & 0.7 & 87.2 & 75.5 & 7.4 & 7.8 & 75.4 \\
 & MegaDes. & 0.7 & 77.2 & 46.0 & 9.5 & 11.5 & 45.6 \\
\midrule
\multirow{2}{*}{StripeSpotter~\cite{lahiri2011biometric}} & MiewID & 3.0 & 94.7 & 94.7 & 3.2 & 0.2 & 94.6 \\
 & MegaDes. & 3.0 & 99.9 & 99.9 & 4.2 & 1.2 & 99.9 \\
\bottomrule
\end{tabular}

\caption{\textbf{Counterfactual foreground retrieval performance.}
Re-evaluating MiewID and MegaDescriptor on selected representative datasets shows that full-image Retrieval@1 often hides qualitatively different evidence profiles. Some datasets are cleanly foreground-supported, some are overdetermined by both animal and shortcut evidence, and others show weak foreground support despite high shortcut recoverability. See full results in Tab.~\ref{tab:supp-all-cfr-components}.}
\label{tab:all-cfr-miewid-megad}
\vspace{-1cm}
\end{table}

\medskip
\noindent\textbf{Shortcut signal does not always lead to method failure.}
High SSI should not be interpreted as a direct failure of the model. In datasets such as OpenCows2020 and AAUZebraFish, CFR remains high. This means that the model can retrieve the correct individual from the animal foreground, even though shortcut-only inputs are also predictive. The issue is therefore not simply that the method ignores the animal; rather, the benchmark contains additional identity-correlated evidence.
A concerning pattern appears when high SSI is paired with low CFR. Saimaa seal and NDD20 show modest full-image retrieval, weaker foreground-only retrieval, and high shortcut recoverability. In this case, full-image Retrieval@1 is weak evidence of animal-foreground identification, because much of the recoverable signal is also present in shortcut-only views. This is the case where standard evaluation is most misleading.

\begin{figure}[b]
\vspace{-3mm}
\centering
\includegraphics[width=0.95\linewidth]{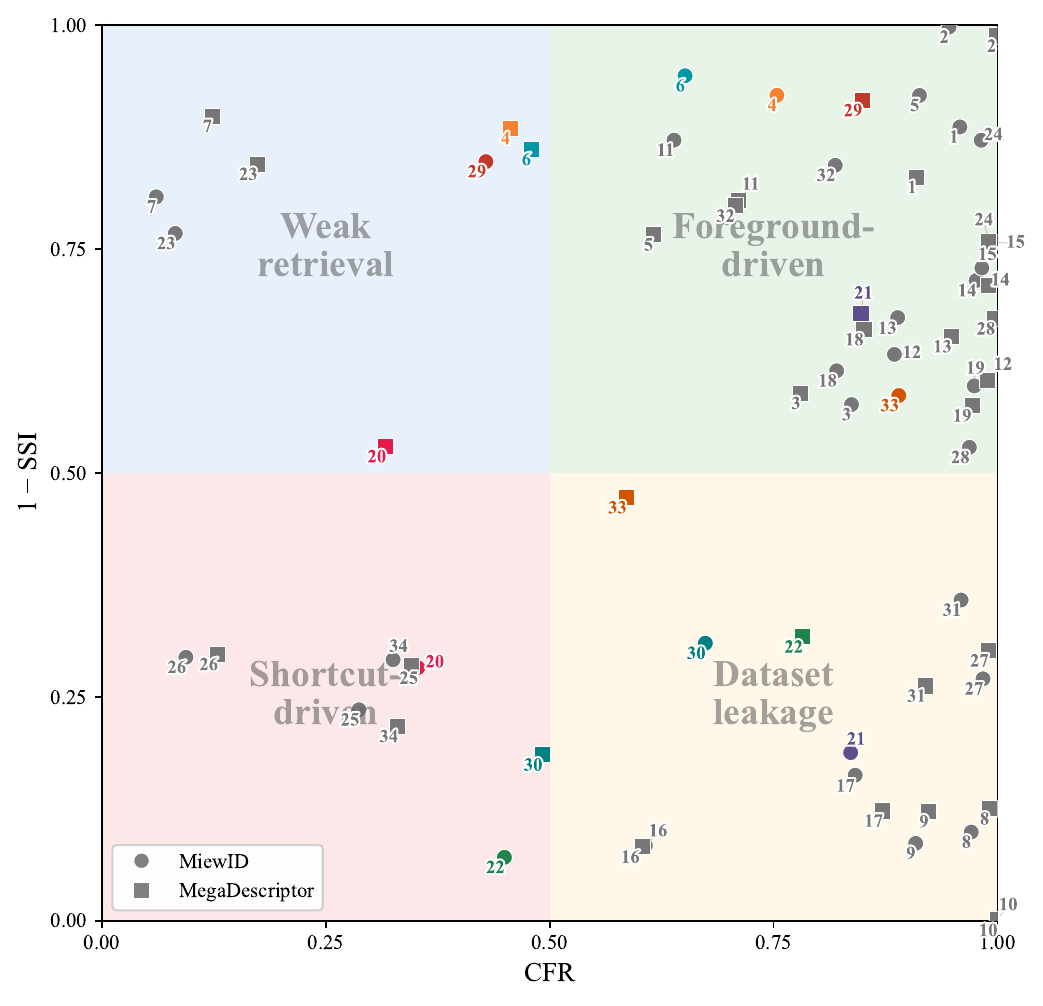}
\vspace{-1mm}
\caption{\textbf{CFR vs. SSI}.
Seemingly similarly performing models can use different types of evidence (colored samples).
Numbered points correspond to the dataset indices in Tab.~\ref{tab:supp-all-cfr-components}.
}
\label{fig:dataset_evidence}
\end{figure}

\begin{figure*}[t]
\centering
\includegraphics[width=0.95\textwidth]{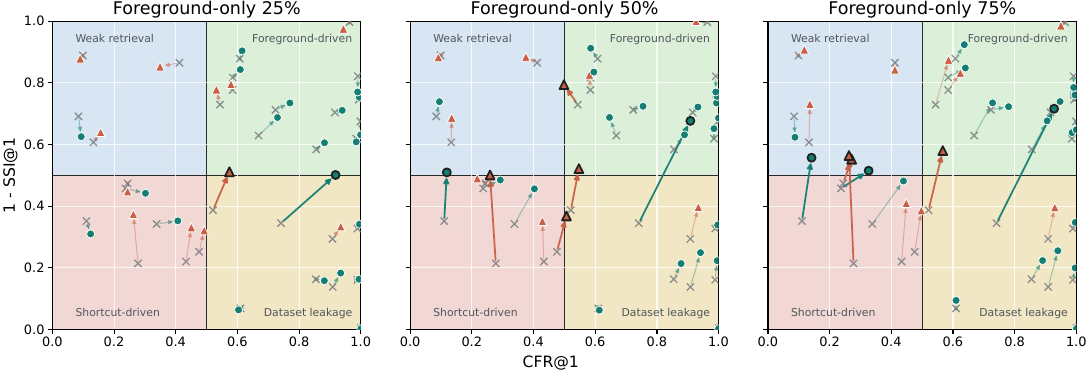}
\caption{\textbf{Evidence-profile changes under foreground-mixed training.}
For each of the 34 datasets, {\color{gray}$\times$} marks the full-image
baseline and its arrow ends at the corresponding foreground-mixed checkpoint.
{\color{teal}$\bigcirc$} denotes the 22 represented datasets and
{\color{orange}$\triangle$} the 12 unseen datasets. Darker arrows and outlined
endpoints mark transitions across the descriptive $0.5$ boundaries.}
\label{fig:megad-mix-ablation}
\end{figure*}

\medskip
\noindent\textbf{Model comparisons reveal evidence differences.}
CFR also shows that models with similar full-image performance can rely on different evidence profiles (see Fig.~\ref{fig:dataset_evidence}).
For example, on HumpbackWhaleID, MiewID and MegaDescriptor have similar full-image Retrieval@1, but MiewID has substantially higher shortcut recoverability, while MegaDescriptor retains comparable foreground support with much lower SSI.
The standard metric suggests similar retrieval performance, whereas CFR shows that the supporting evidence differs.
Other datasets show the opposite. On IPanda50 and NDD20, MegaDescriptor obtains much higher full-image Retrieval@1 than MiewID, but this improvement is only partly reflected in CFR and is accompanied by high shortcut recoverability. These cases show why model comparison should not be based only on full-image retrieval: an apparent performance gain may improve foreground or shortcut recoverability, or both.

\medskip
\noindent\textbf{Low shortcut signal gives cleaner interpretation.}
Datasets with low SSI provide the cleanest interpretation of standard retrieval performance. StripeSpotter, SeaTurtleID2022, Reunion turtles, Amvrakikos turtles, WhaleSharkID, BelugaID, and MacaqueFaces have low shortcut recoverability relative to their full-image performance. On these datasets, foreground-only retrieval largely preserves the useful signal, while shortcut-only retrieval contributes little. Full-image Retrieval@1 is therefore more directly supported by animal-visible evidence.
Low shortcut signal should not be confused with high retrieval performance. NyalaData has low SSI, but MegaDescriptor has substantially lower CFR than MiewID. This shows that low SSI only indicates that the observed signal is not primarily recoverable from shortcut-only inputs; it does not imply that the method retrieves individuals accurately.

\medskip
\noindent\textbf{Takeaway.} Our evaluation shows that full-image Retrieval@$k$ is not sufficient to interpret animal identification results. Similar scores can arise from foreground-supported matching, overdetermined benchmarks where both animal and shortcut evidence are predictive, or cases where foreground support is weak. CFR and SSI keep the standard retrieval task intact, but turn a single performance number into an evidence profile that shows whether the retrieved identity is supported by the animal foreground and/or shortcut.

\section{Foreground-Mixed Training as Mitigation}
\label{sec:foreground-mix-ablation}

One possible explanation for low foreground retrieval is that foreground-only
inputs are unfamiliar to models trained only on full images. If so, exposing a
model to masked inputs during training should improve CFR and reduce SSI. If
shortcut cues remain predictive after this intervention, they are more likely
to reflect properties of the data than the foreground-only input itself.

\medskip
\noindent\textbf{Experimental setup.}
We compare four MegaDescriptor-L-384 checkpoints trained on the same 22-dataset
collection. The reference checkpoint, MegaD$^C$, uses only full images. The
other checkpoints use a foreground-only input with probability $0.25$, $0.50$,
or $0.75$, and the full image otherwise. If a selected sample has no accepted
mask, we retain its full image. All checkpoints are evaluated using the same
accepted annotations and CFR protocol. We report results for the 22 represented
datasets and 12 unseen datasets to distinguish adaptation from transfer.

\medskip
\noindent\textbf{Results.}
Foreground mixing generally lowers shortcut recoverability, with clearer
changes at $50\%$ and $75\%$ exposure (Fig.~\ref{fig:megad-mix-ablation}). The
number of shortcut-driven datasets is unchanged at $25\%$, but falls from seven
under the baseline to four at $50\%$ and three at $75\%$; no mixture moves a
dataset into that regime. IPanda50 moves from the high-CFR, high-SSI region to
foreground-supported retrieval, while HyenaID2022 makes the same transition
despite being unseen during training.
The intervention is not a complete solution. Some datasets leave the
shortcut-driven regime because SSI decreases while CFR remains low, indicating
weaker shortcut recoverability rather than strong foreground recognition.
AerialCattle2017 remains high in both CFR and SSI at every exposure level,
showing that foreground-mixed training does not remove all identity-correlated
dataset cues. Thus, unfamiliar masked inputs explain part, but not all, of the
foreground deficit. CFR and SSI can therefore guide mitigation while also
revealing when the remaining issue lies in the data.

\section{Discussion}

One of the main goals of evaluation is to guide future development. A benchmark score tells us whether a model performs well, but it rarely tells us where to spend the next six months of work. Should we build a better model? Collect more data? Or rethink the benchmark itself? Counterfactual retrieval addresses this gap by asking not only whether retrieval was correct, but also what visual evidence supported it. By separating foreground-supported retrieval from shortcut-supported retrieval, it provides information that standard retrieval metrics cannot, making it easier to decide what to investigate next.

\textbf{The evidence profile, however, is only the starting point. By itself, it does not tell us what to change.} Instead, it turns a single benchmark score into a more focused set of questions. Is the model struggling to recognize individuals? Is the benchmark making the task artificially easy? Or are both contributing? Different answers naturally lead to different solutions. Weak foreground support points toward the recognition model, training data, or species-specific representations. Strong shortcut recoverability shifts attention to the benchmark itself, raising questions about data split, repeated backgrounds, acquisition conditions, or other identity-correlated cues. In many cases, both the model and the benchmark deserve closer attention.

\textbf{In any case, answering these questions requires looking beyond any metric.} Inspecting representative examples, reviewing dataset metadata, examining train/test splits, or using feature-level interpretability methods such as Grad-CAM~\cite{selvaraju2017grad} and attention visualization~\cite{caron2021emerging} can help identify where the recovered evidence comes from. Together, these analyses can reveal whether the observed behavior reflects shortcut cues, benchmark design, or genuine visual representations. We see these methods as complementary. Counterfactual retrieval highlights \textit{that shortcut exists} at the benchmark level, while feature-level analysis helps explain \textit{where it comes} from and why.

\section{Conclusion}
\label{sec:conclusion}

Standard retrieval metrics tell us whether the correct individual was retrieved, but not what evidence made that retrieval possible. With that in mind, we introduced Counterfactual Foreground Retrieval that make the evaluation evidence measurable rather than implicit.
Across a wide range of datasets, we show that similar retrieval performance can reflect fundamentally different evidence profiles. Rather than introducing another score to optimize, CFR and SSI provide information that helps guide model development, benchmark design, and future investigation. We hope this encourages evaluation that asks not only whether retrieval is accurate, but also what evidence makes it accurate.

\textbf{However, evidence-aware evaluation is only one part of this process.} CFR is a diagnostic rather than a causal explanation of model behavior. It reveals whether retrieval remains supported by animal-visible evidence under the proposed counterfactual protocol, but it does not identify the exact source of shortcut cues or explain why they arise. Understanding those mechanisms still benefits from complementary analyses, such as feature-level interpretability, dataset inspection, and ecological expertise. We see these directions as natural next steps toward more reliable and interpretable wildlife identification systems.

\section*{Acknowledgments}
This work was supported by the Imageomics Institute through the U.S. National
Science Foundation's Harnessing the Data Revolution program (Award
No.~2118240). We also acknowledge the Ohio Supercomputer Center for providing
the computational and storage resources used in this study.

{
    \small
    \bibliographystyle{ieeenat_fullname}
    \bibliography{main}
}

\appendix

\definecolor{wacvblue}{rgb}{0.21,0.49,0.74}

\def\wacvPaperID{1869}
\def\confName{WACV}
\def\confYear{2027}

\renewcommand{\thesection}{S\arabic{section}}
\renewcommand{\thetable}{S\arabic{table}}
\renewcommand{\thefigure}{S\arabic{figure}}

\maketitlesupplementary
\setcounter{section}{0}
\setcounter{table}{0}
\setcounter{figure}{0}

\section{Masks and Counterfactual Inputs}
\label{app:mask-processing}
\label{app:mask-coverage}

\noindent\textbf{Mask generation details}: Given that the available datasets span diverse species and annotations, mask generation cannot simply rely on a single prompt and strategy.
We therefore generated SAM3~\cite{kirillov2023segment,sam3}-mask using dataset-specific prompts matched to the species and dataset.
Depending on the dataset, prompts referred either to the whole animal (e.g., ``cow'' and ``fish'') or to the annotated body part (e.g., ``sea turtle head'' and ``bear head''). To get the best mask possible, we also included alternative prompts, when appropriate, and selected the highest-confidence one. For more details about the prompts used, see Table~\ref{tab:sam-mask-coverage}.
For datasets with available bounding boxes, we compared all SAM3 boxes with the annotation and selected the one with the highest IoU. We discarded all masks with IoU $<$ 0.5.

\begin{table}[!b]
\vspace{-4mm}
\centering
\footnotesize
\setlength{\tabcolsep}{3pt}
\begin{tabular}{@{}p{0.16\textwidth}p{0.2\textwidth}rr@{}}
\toprule
\textbf{Dataset} & \textbf{Prompt(s)} & \textbf{Masks} & \textbf{[\%]} \\
\midrule
AAUZebraFish~\cite{bruslund2020re} & zebrafish, fish & 6,671 & 100 \\
ATRW~\cite{li2019atrw} & tiger & 5,398 & 99.7 \\
AerialCattle2017~\cite{andrew2017visual} & cow, cattle & 30,010 & 64.8 \\
Amvrakikos turtles~\cite{adam2025exploiting} & sea turtle head, sea turtle, turtle & 195 & 97.5 \\
BelugaID~\cite{belugaid2022} & beluga whale, whale & 4,916 & 83.3 \\
BirdIndividualID~\cite{ferreira2020deep} & bird & 52,076 & 98.7 \\
BrownBearHeads~\cite{rosenberg2026individual} & bear head, brown bear, bear & 72,930 & 100 \\
CTai~\cite{freytag2016chimpanzee} & chimpanzee & 5,043 & 99.3 \\
CZoo~\cite{freytag2016chimpanzee} & chimpanzee & 2,105 & 99.8 \\
Cows2021~\cite{gao2021towards} & cow, cattle & 7,915 & 91.3 \\
CzechLynx~\cite{picek2026czechlynx} & lynx, Eurasian lynx & 33,858 & 85.2 \\
FriesianCattle2015~\cite{andrew2016automatic} & cow, cattle & 325 & 86.2 \\
FriesianCattle2017~\cite{andrew2017visual} & cow, cattle & 938 & 99.8 \\
GiraffeZebraID~\cite{parham2017animal} & giraffe, zebra & 6,917 & 99.9 \\
Giraffes~\cite{miele2021revisiting} & giraffe & 1,392 & 99.9 \\
HappyWhale~\cite{kaggle2022happywhale} & whale & 78,851 & 99.8 \\
HumpbackWhaleID~\cite{kaggle2019humpback} & humpback whale, whale & 33,156 & 99.5 \\
HyenaID2022~\cite{hyenaid2022} & hyena & 2,992 & 95.6 \\
IPanda50~\cite{wang2021giant} & giant panda, panda & 6,874 & 100 \\
LeopardID2022~\cite{leopardid2022} & leopard & 4,516 & 98.9 \\
LionData~\cite{dlamini2020automated} & lion & 705 & 95.3 \\
MacaqueFaces~\cite{witham2018automated} & macaque, monkey & 6,280 & 100 \\
Melops~\cite{sordalen2026wild} & fish & 24,578 & 100 \\
NDD20~\cite{trotter2020ndd20} & dolphin & 6,090 & 99.8 \\
NOAARightWhale~\cite{kaggle2015rightwhale} & right whale, whale & 11,437 & 99.7 \\
NyalaData~\cite{dlamini2020automated} & animal & 1,942 & 100 \\
OpenCows2020~\cite{andrew2021visual} & cow, cattle & 3,276 & 69.2 \\
Reunion turtles~\cite{adam2025exploiting} & sea turtle, turtle & 336 & 100 \\
SMALST~\cite{zuffi2019three} & zebra & 12,808 & 99.7 \\
Saimaa seal~\cite{nepovinnykh2022sealid} & seal & 2,850 & 97.1 \\
SeaTurtleID2022~\cite{adam2024seaturtleid2022} & sea turtle head, sea turtle & 8,143 & 95.5 \\
StripeSpotter~\cite{lahiri2011biometric} & zebra & 824 & 100 \\
WhaleSharkID~\cite{holmberg2009estimating} & whale shark & 7,603 & 98.8 \\
ZindiTurtleRecall~\cite{zindi2022turtlerecall} & sea turtle, turtle & 13,288 & 100 \\
\midrule
\textbf{Total} & -- & \textbf{457,238} & \textbf{94.3} \\
\bottomrule
\end{tabular}
\caption{\textbf{SAM3-assisted mask extraction details.} Prompts used for each dataset, together with the number of accepted masks.}
\label{tab:sam-mask-coverage}
\end{table}

\noindent\textbf{Mask quality assurance details}: To further assess mask quality, we manually annotated the visible foreground (i.e., the body of the animal) for 50 randomly sampled images from eight representative datasets using X-AnyLabeling~\cite{X-AnyLabeling} and compared the resulting masks with the corresponding SAM3 predictions using IoU. Median IoU exceeded 96\% for every audited dataset and reached 98.2\% overall (Table~\ref{tab:mask_quality_audit}). The main failure mode occurred in NyalaData, where the generic ``animal’’ prompt occasionally selected the wrong individual in multi-animal scenes.
Overall, the audited masks are highly accurate, with errors substantially smaller than the controlled low- and medium-severity mask perturbations under which CFR and SSI remain stable (Sec.~\ref{app:cfr-robustness}). This more than suggests that the observed mask quality is well within the range needed for reliable CFR evaluation.

\medskip
\noindent\textbf{Making counterfactual inputs}:
All accepted masks were used to construct the counterfactual inputs evaluated by CFR. All views were generated from the original annotation crop used for the standard retrieval evaluation. The full-image view was left unchanged. The foreground-only view retained pixels inside the accepted mask, whereas the shortcut-only view retained those outside it. The silhouette-only view replaced the foreground texture with a solid black silhouette. All removed regions were filled with the ImageNet RGB mean before applying each model's standard resizing and normalization. Consequently, every view shared the same image crop, identity labels, query--gallery protocol, and preprocessing, differing only in the visual evidence available to the retrieval model. To allow reproducibility and future use,
we made \href{https://github.com/Imageomics/cfr}{the code} publicly available.

\begin{table}[h]
\centering
\small
\setlength{\tabcolsep}{2.5pt}
\begin{tabular}{lrrr}
\toprule
        & Median IoU & Mean IoU & IoU $>90\%$ \\
Dataset & (\%) & (\%) & (\%) \\
\midrule
BrownBearHeads~\cite{rosenberg2026individual} & 99.3 & 98.9 & 100.0 \\
CzechLynx~\cite{picek2026czechlynx} & 98.3 & 97.6 & 94.0 \\
HyenaID2022~\cite{hyenaid2022} & 97.9 & 97.2 & 96.0 \\
LeopardID2022~\cite{leopardid2022} & 96.2 & 95.1 & 92.0 \\
NyalaData~\cite{dlamini2020automated} & 97.0 & 87.3 & 90.0 \\
Saimaa seal & 98.9 & 97.6 & 96.0 \\
SeaTurtleID2022~\cite{adam2024seaturtleid2022} & 97.2 & 95.1 & 92.0 \\
WhaleSharkID~\cite{holmberg2009estimating} & 97.8 & 96.2 & 96.0 \\
\midrule
Overall & 98.2 & 96.1 & 95.1 \\
\bottomrule
\end{tabular}
\caption{Manual mask-quality audit of SAM masks. IoU $>90\%$ reports the percentage of manual reference masks whose SAM mask exceeds 90\% IoU.}
\label{tab:mask_quality_audit}
\end{table}

\section{Sensitivity to Mask and Image Corruptions}
\label{app:cfr-robustness}

CFR and SSI rely on counterfactual image views generated from segmentation masks.
Their practical usefulness therefore depends on whether these evidence profiles remain stable under realistic annotation errors and common image degradations. We evaluate this directly by applying controlled perturbations to the masks and images while keeping the retrieval task, annotations, gallery, and models fixed.
We consider nine deterministic perturbations at three severity levels across all 34 datasets and both evaluated models, resulting in $3\times9\times34\times2=1836$ paired evaluations.

\medskip
\noindent\textbf{Experimental setup.}
We consider four mask perturbations and five image perturbations. Mask perturbations modify approximately $5\%$, $10\%$, or $20\%$ of the accepted foreground area by applying erosion, dilation, interior false negatives, or exterior false positives. Image perturbations include Gaussian noise, Gaussian blur, JPEG compression, foreground CutOut, and background CutOut at low, medium, and high severity levels.

\begin{figure}[b]
    \centering
    \includegraphics[width=\linewidth]{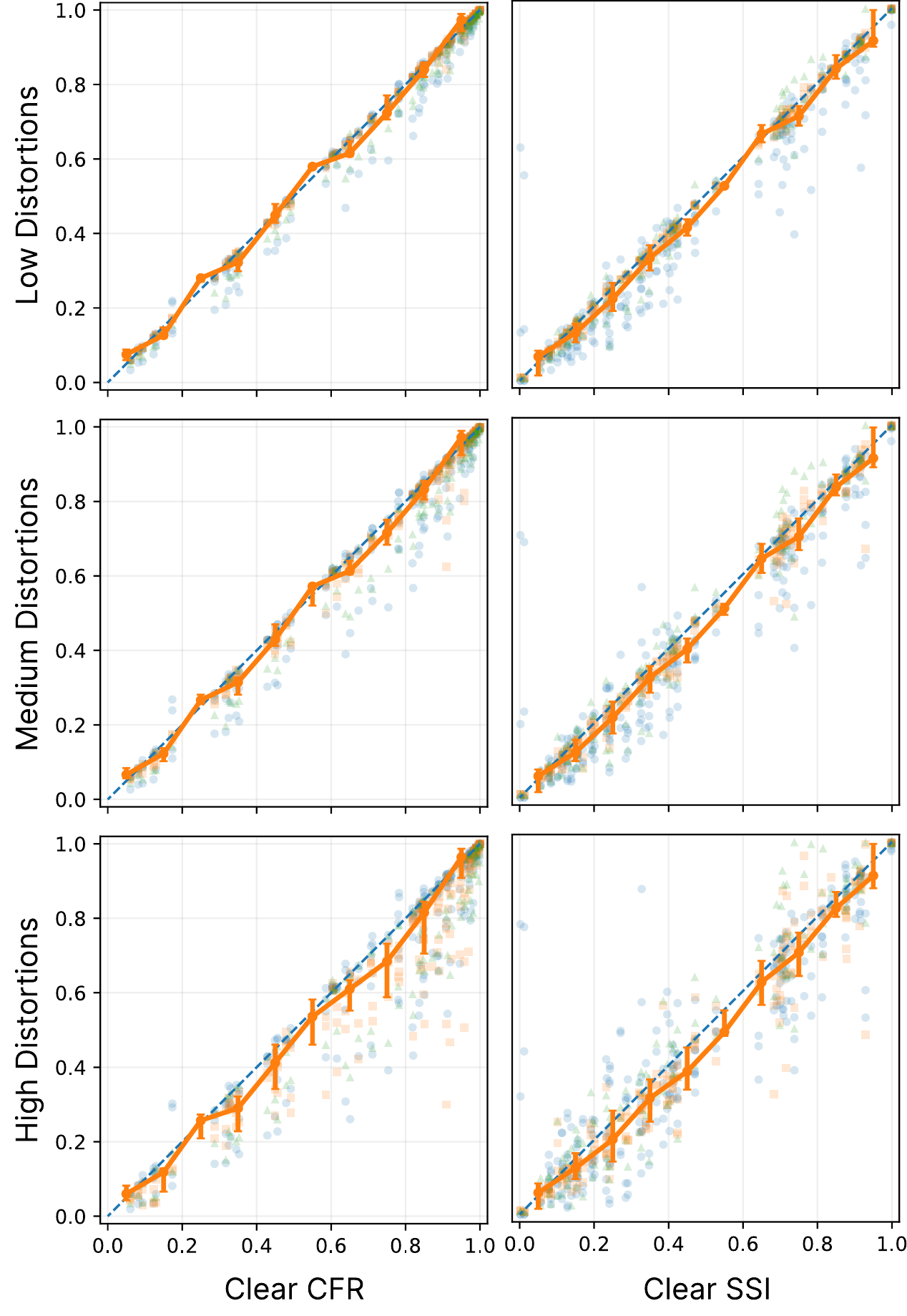}
    \caption{\textbf{Robustness of CFR and SSI to controlled mask and image perturbations.}  CFR and SSI remain stable under moderate perturbations and change primarily when discriminative animal evidence is removed, consistent with their intended interpretation. Each point represents one paired evaluation on 34 datasets using two models (e.g., MegaDescriptor and MiewID), comparing the clean metric with its perturbed counterpart. Rows correspond to increasing perturbation severity (low, medium, high), and columns show Clear CFR and SSI. {\color{blue}$\bigcirc$} indicates mask perturbations, {\color{orange}$\square$} generic image degradations, and {\color{green}$\triangle$} targeted evidence removal (e.g., foreground / background CutOut). The orange curve shows the median perturbed value in ten equal-width bins of the clean metric, with interquartile-range error bars.}
    \label{fig:scatterplot}
\end{figure}

\medskip
\noindent\textbf{Boundary errors have little effect.} Across all datasets, both metrics remain close to the identity line under low and medium perturbations (Fig.~\ref{fig:scatterplot}), indicating that moderate segmentation inaccuracies have little influence on the resulting evidence profiles. Boundary erosion, dilation, and exterior false positives consistently produce only small changes, even at the highest severity. These results suggest that CFR and SSI are robust to the level of mask imprecision expected from modern automatic segmentation models.

\medskip
\noindent\textbf{Removing animal parts changes the metrics as expected.}
The largest mask-induced changes occur for interior false negatives, which remove regions from the animal itself rather than perturbing its boundary. Likewise, among image perturbations, Foreground CutOut consistently produces larger changes than Background CutOut. This behavior matches the intended interpretation of both metrics. CFR decreases when discriminative foreground appearance is removed, while SSI changes according to how evidence is redistributed between the foreground and shortcut views. In contrast, Background CutOut removes contextual information while leaving the animal appearance intact, producing substantially smaller changes.

\medskip
\noindent\textbf{Common image degradations have limited impact.} Gaussian noise, blur, and JPEG compression have little effect at low and medium severity and produce noticeable changes only under strong degradation. Importantly, these perturbations affect the entire image rather than selectively removing identity evidence, making their impact substantially smaller than Foreground CutOut. This indicates that the proposed evidence profiles are more sensitive to the loss of discriminative animal information than to typical image-quality degradation.

\medskip

\noindent\textbf{Overall evaluation.}
Our experiments show that the conclusions drawn from CFR and SSI are largely insensitive to realistic imperfections in the counterfactual construction. Moderate mask errors and common image degradations have little or no influence on the resulting evidence profiles, whereas perturbations that remove discriminative animal appearance produce the expected changes in both metrics. This gives us confidence that CFR and SSI capture meaningful differences in the visual evidence supporting retrieval rather than artifacts from imperfect masks.

\begin{table*}[!h]
\centering
\scriptsize
\setlength{\tabcolsep}{3pt}
\renewcommand{\arraystretch}{1.1}
\begin{tabular}{@{}p{0.14\textwidth}p{0.20\textwidth}p{0.20\textwidth}p{0.1\textwidth}p{0.21\textwidth}p{0.09\textwidth}@{}}
\toprule
\textbf{Dataset} & \textbf{Species or group} & \textbf{Image conditions} & \textbf{CFR unit} & \textbf{Intended biometric} & \textbf{Silhouette role} \\
\midrule
AAUZebraFish~\cite{bruslund2020re} & Zebrafish (teleost fish) & Controlled tank video & Image/crop & Stripe pattern, color, morphology & Ambiguous \\
ATRW~\cite{li2019atrw} & Amur tiger (felid) & Zoo video, varied pose & Image/crop & Coat stripe pattern & Shortcut \\
AerialCattle2017~\cite{andrew2017visual} & Holstein-Friesian cattle (bovid) & Outdoor drone video & Image/crop & Coat pattern & Ambiguous \\
Amvrakikos turtles~\cite{adam2025exploiting} & Loggerhead turtle (reptile) & Boat-based head photographs & Head bbox & Facial scale pattern & Shortcut \\
BelugaID~\cite{belugaid2022} & Beluga whale (cetacean) & Aerial/field top-view images & Image/crop & Scars, pigmentation, dorsal ridge & Ambiguous \\
BirdIndividualID~\cite{ferreira2020deep} & Three bird species & Feeder cameras, captive images & Image/crop & Plumage and facial appearance & Ambiguous \\
BrownBearHeads~\cite{rosenberg2026individual} & Brown bear (ursid) & Longitudinal wild head photos & Head crop & Facial appearance, head morphology & Ambiguous \\
CTai~\cite{freytag2016chimpanzee} & Chimpanzee (primate) & Face recordings, variable quality & Face crop & Facial appearance and geometry & Ambiguous \\
CZoo~\cite{freytag2016chimpanzee} & Chimpanzee (primate) & Captive high-quality face images & Face crop & Facial appearance and geometry & Ambiguous \\
Cows2021~\cite{gao2021towards} & Holstein-Friesian cattle (bovid) & Fixed overhead farm video & Image/crop & Coat pattern & Ambiguous \\
CzechLynx~\cite{picek2026czechlynx} & Eurasian lynx (felid) & Long-term camera-trap images & Image/crop & Coat and spot pattern & Shortcut \\
FriesianCattle2015~\cite{andrew2016automatic} & Holstein-Friesian cattle (bovid) & Overhead milking-area video & Image/crop & Coat pattern & Ambiguous \\
FriesianCattle2017~\cite{andrew2017visual} & Holstein-Friesian cattle (bovid) & Overhead milking-area video & Image/crop & Coat pattern & Ambiguous \\
GiraffeZebraID~\cite{parham2017animal} & Masai giraffe \& plains zebra & Citizen-science field census & Image/crop & Coat or stripe pattern & Shortcut \\
Giraffes~\cite{miele2021revisiting} & Giraffe (giraffid) & Wild field photographs & Image/crop & Coat pattern & Shortcut \\
HappyWhale~\cite{kaggle2022happywhale} & Multiple cetaceans & Marine survey images & Image/crop & Fin or fluke contour, scars & Identity \\
HumpbackWhaleID~\cite{kaggle2019humpback} & Humpback whale (cetacean) & Tail flukes above water & Image/crop & Fluke contour and pigmentation & Identity \\
HyenaID2022~\cite{hyenaid2022} & Spotted hyena (hyaenid) & Wild day and night photographs & Body bbox & Spot pattern and scars & Shortcut \\
IPanda50~\cite{wang2021giant} & Giant panda (ursid) & Captive streaming-video frames & Image/crop & Facial and coat appearance & Ambiguous \\
LeopardID2022~\cite{leopardid2022} & Leopard (felid) & Wild day and night photographs & Body bbox & Rosette pattern and scars & Shortcut \\
LionData~\cite{dlamini2020automated} & Lion (felid) & Wild photos of body parts & Part crop & Ear, nose, facial markings & Ambiguous \\
MacaqueFaces~\cite{witham2018automated} & Rhesus macaque (primate) & Indoor facility video & Face crop & Facial appearance and geometry & Ambiguous \\
Melops~\cite{sordalen2026wild} & Corkwing wrasse (teleost fish) & Capture-recapture side photographs & Head/body crop & Asymmetric facial color pattern & Shortcut \\
NDD20~\cite{trotter2020ndd20} & Two dolphin species (cetacean) & Boat photos and underwater video & Image/crop & Dorsal-fin contour, markings & Identity \\
NOAARightWhale~\cite{kaggle2015rightwhale} & Right whale (cetacean) & Distant aerial survey images & Image/crop & Callosity pattern and scars & Ambiguous \\
NyalaData~\cite{dlamini2020automated} & Nyala (bovid) & Uniform wild whole-body images & Image/crop & Coat stripes, markings, scars & Shortcut \\
OpenCows2020~\cite{andrew2021visual} & Holstein-Friesian cattle (bovid) & Combined farm video & Image/crop & Coat pattern & Ambiguous \\
Reunion turtles~\cite{adam2025exploiting} & Green \& hawksbill turtles (reptile) & Citizen-diver head profiles & Head crop & Facial scale pattern & Shortcut \\
SMALST~\cite{zuffi2019three} & Grevy's zebra (synthetic) & Synthetic 3D animals & Rendered crop & Stripe texture and geometry & Ambiguous \\
Saimaa seal~\cite{nepovinnykh2022sealid} & Saimaa ringed seal (pinniped) & Boat and game-camera field images & Crop; some bbox & Fur-ring pelage pattern & Shortcut \\
SeaTurtleID2022~\cite{adam2024seaturtleid2022} & Loggerhead turtle (reptile) & Long-term underwater field photos & Head bbox & Lateral/dorsal head-scale pattern & Shortcut \\
StripeSpotter~\cite{lahiri2011biometric} & Plains zebra (equid) & Vehicle-based field photography & Flank region & Flank stripe pattern & Shortcut \\
WhaleSharkID~\cite{holmberg2009estimating} & Whale shark (elasmobranch) & Underwater marine survey photos & Body bbox & Spot pattern and scars & Shortcut \\
ZindiTurtleRecall~\cite{zindi2022turtlerecall} & Green turtle (reptile) & Rehabilitation and field close-ups & Eye/scute crop & Post-ocular scute pattern & Shortcut \\
\bottomrule
\end{tabular}
\caption{\textbf{Characteristics of the CFR benchmark datasets.}
A systematic summary of the evaluation datasets, complementing the mask-coverage
details in Table~\ref{tab:sam-mask-coverage}. \emph{CFR unit} is the localized
region actually embedded (whole image, body/head crop, or bounding box), not
every annotation released with the source dataset. \emph{Intended biometric} is
the animal-intrinsic evidence each dataset targets; the foreground-only view is
the input, and identity is carried by this evidence inside it. \emph{Silhouette
role} records whether body outline is a \emph{shortcut} (outline is not the
intended biometric), \emph{identity} (outline is the biometric, e.g., a fin or
fluke contour), or \emph{ambiguous} (both readings are plausible); it sets how
silhouette enters $S_{\mathrm{shortcut}}$ and SSI (Sec.~\ref{sec:cfr-signal}).
Descriptions follow the WildlifeDatasets overview~\cite{vcermak2024wildlifedatasets}.}
\label{tab:dataset-char}
\end{table*}

\section{Other Characteristics and Silhouette Roles}
\label{app:dataset-characteristics}

The CFR benchmark spans 34 datasets with very different species, acquisition setups, and annotation types. Table~\ref{tab:dataset-char} summarizes each by species or group, image conditions, the CFR unit we embed, the intended biometric, and the silhouette role. They cover fish, reptiles, birds, and a range of mammals, under conditions from zoo video to camera traps, drones, aerial surveys, and underwater photography.

\bigskip
\noindent\textbf{Species- and dataset-dependent silhouette role.}
The foreground-only view is the input, not the identity cue: identity is carried by a biometric inside that view, such as a coat or stripe pattern, facial scales, or a fin contour, which the \emph{Intended biometric} column names. Whether body outline is a shortcut or legitimate evidence depends on the species. For animals identified by an outline, such as a dorsal fin or whale fluke, silhouette is part of the intended biometric; for animals identified by a surface pattern it is a shortcut. We label each dataset \emph{shortcut}, \emph{identity}, or \emph{ambiguous}, and this label sets how silhouette enters the shortcut signal (Sec.~\ref{sec:cfr-signal}): shortcut datasets include it through $S_{\mathrm{shortcut}}=\max(S_{T_{\mathrm{sc}}}@k, S_{T_{\mathrm{sil}}}@k)$, identity datasets use only $S_{T_{\mathrm{sc}}}@k$ and report silhouette separately, and ambiguous datasets are reported under both definitions with the choice fixed before results are seen. The labels describe the intended biometric, not a model's internal features.

\bigskip
\noindent\textbf{The shortcut view retains a negative silhouette.}
The texture-removed shortcut view $T_{\mathrm{sc}}$ replaces animal pixels with a
neutral fill but keeps the animal-shaped region in place, so it measures scene
context together with a negative silhouette, not pure background. A low SSI
therefore narrows an outline-only or context-only explanation without fully
removing outline information from the shortcut view. This is also why silhouette
is scored as a separate view $T_{\mathrm{sil}}$ rather than assumed absent from
$T_{\mathrm{sc}}$.

\begin{minipage}{\textwidth}
\centering
\scriptsize
\setlength{\tabcolsep}{1.8pt}
\renewcommand{\arraystretch}{0.78}
\resizebox{0.95\textwidth}{!}{%
\begin{tabular}[t]{@{}rllrrrrrrr@{}}
\toprule
\textbf{\#} & \textbf{Dataset} & \textbf{Model} & \textbf{Rand.} & \textbf{Full} & \textbf{Frg.} & \textbf{SC} & \textbf{Sil} & \textbf{SSI} & \textbf{CFR} \\
\midrule
\multirow{2}{*}{1} & \multirow{2}{*}{SeaTurtleID2022} & MiewID & \multirow{2}{*}{0.5} & 96.9 & 95.8 & 11.4 & 5.7 & 11.4 & 95.8 \\
 & & MegaD & & 92.4 & 91.0 & 16.1 & 4.9 & 17.0 & 91.0 \\
\midrule
\multirow{2}{*}{2} & \multirow{2}{*}{StripeSpotter} & MiewID & \multirow{2}{*}{3.0} & 94.7 & 94.7 & 3.2 & 3.2 & 0.2 & 94.6 \\
 & & MegaD & & 99.9 & 99.9 & 4.2 & 3.2 & 1.2 & 99.9 \\
\midrule
\multirow{2}{*}{3} & \multirow{2}{*}{HyenaID2022} & MiewID & \multirow{2}{*}{0.7} & 87.4 & 83.8 & 37.4 & 22.0 & 42.4 & 83.7 \\
 & & MegaD & & 92.5 & 78.2 & 38.5 & 16.1 & 41.1 & 78.0 \\
\midrule
\multirow{2}{*}{4} & \multirow{2}{*}{NyalaData} & MiewID & \multirow{2}{*}{0.7} & 87.2 & 75.5 & 7.4 & 5.3 & 7.8 & 75.4 \\
 & & MegaD & & 77.2 & 46.0 & 9.5 & 3.3 & 11.5 & 45.6 \\
\midrule
\multirow{2}{*}{5} & \multirow{2}{*}{Reunion turtles} & MiewID & \multirow{2}{*}{0.9} & 92.0 & 91.4 & 8.0 & 5.7 & 7.8 & 91.3 \\
 & & MegaD & & 61.9 & 64.0 & 15.2 & 6.0 & 23.4 & 61.6 \\
\midrule
\multirow{2}{*}{6} & \multirow{2}{*}{Amvrakikos turtles} & MiewID & \multirow{2}{*}{1.5} & 65.6 & 70.8 & 5.1 & 3.1 & 5.6 & 65.1 \\
 & & MegaD & & 49.7 & 48.7 & 8.2 & 2.6 & 13.9 & 47.9 \\
\midrule
\multirow{2}{*}{7} & \multirow{2}{*}{Melops} & MiewID & \multirow{2}{*}{0.0} & 6.3 & 6.1 & 1.2 & 1.0 & 19.2 & 6.1 \\
 & & MegaD & & 12.9 & 12.3 & 1.3 & 0.6 & 10.2 & 12.3 \\
\midrule
\multirow{2}{*}{8} & \multirow{2}{*}{AAUZebraFish} & MiewID & \multirow{2}{*}{16.7} & 98.3 & 97.6 & 90.2 & 88.9 & 90.1 & 97.1 \\
 & & MegaD & & 100.0 & 99.3 & 89.6 & 84.7 & 87.5 & 99.1 \\
\midrule
\multirow{2}{*}{9} & \multirow{2}{*}{ATRW} & MiewID & \multirow{2}{*}{1.2} & 94.1 & 91.0 & 86.0 & 65.8 & 91.4 & 90.9 \\
 & & MegaD & & 95.9 & 92.4 & 84.4 & 54.7 & 87.8 & 92.3 \\
\midrule
\multirow{2}{*}{10} & \multirow{2}{*}{AerialCattle2017} & MiewID & \multirow{2}{*}{7.9} & 100.0 & 100.0 & 99.9 & 99.8 & 99.9 & 100.0 \\
 & & MegaD & & 100.0 & 100.0 & 99.8 & 99.7 & 99.8 & 100.0 \\
\midrule
\multirow{2}{*}{11} & \multirow{2}{*}{BelugaID} & MiewID & \multirow{2}{*}{0.4} & 77.4 & 64.0 & 10.3 & 5.9 & 12.8 & 63.9 \\
 & & MegaD & & 84.1 & 71.2 & 16.8 & 4.5 & 19.6 & 71.1 \\
\midrule
\multirow{2}{*}{12} & \multirow{2}{*}{BirdIndividualID} & MiewID & \multirow{2}{*}{2.2} & 88.7 & 90.0 & 34.0 & 30.2 & 36.8 & 88.5 \\
 & & MegaD & & 99.1 & 98.9 & 40.7 & 30.7 & 39.6 & 98.8 \\
\midrule
\multirow{2}{*}{13} & \multirow{2}{*}{CTai} & MiewID & \multirow{2}{*}{2.7} & 94.0 & 89.2 & 32.5 & 18.9 & 32.6 & 88.9 \\
 & & MegaD & & 98.0 & 95.0 & 35.9 & 16.4 & 34.8 & 94.9 \\
\midrule
\multirow{2}{*}{14} & \multirow{2}{*}{CZoo} & MiewID & \multirow{2}{*}{4.2} & 98.1 & 97.7 & 31.0 & 22.9 & 28.5 & 97.6 \\
 & & MegaD & & 99.9 & 99.0 & 32.0 & 16.5 & 29.1 & 99.0 \\
\midrule
\multirow{2}{*}{15} & \multirow{2}{*}{Cows2021} & MiewID & \multirow{2}{*}{0.7} & 98.7 & 98.3 & 27.3 & 25.9 & 27.1 & 98.3 \\
 & & MegaD & & 99.8 & 99.7 & 24.7 & 13.5 & 24.2 & 99.7 \\
\midrule
\multirow{2}{*}{16} & \multirow{2}{*}{FriesianCattle2015} & MiewID & \multirow{2}{*}{3.7} & 62.7 & 62.1 & 57.5 & 57.8 & 91.6 & 60.7 \\
 & & MegaD & & 63.4 & 61.8 & 58.4 & 56.2 & 91.7 & 60.4 \\
\midrule
\multirow{2}{*}{17} & \multirow{2}{*}{FriesianCattle2017} & MiewID & \multirow{2}{*}{2.2} & 90.6 & 84.5 & 76.2 & 57.4 & 83.7 & 84.1 \\
 & & MegaD & & 99.2 & 87.5 & 87.4 & 54.9 & 87.7 & 87.2 \\
\bottomrule
\end{tabular}
\hspace{3mm}
\begin{tabular}[t]{@{}rllrrrrrrr@{}}
\toprule
\textbf{\#} & \textbf{Dataset} & \textbf{Model} & \textbf{Rand.} & \textbf{Full} & \textbf{Frg.} & \textbf{SC} & \textbf{Sil} & \textbf{SSI} & \textbf{CFR} \\
\midrule
\multirow{2}{*}{18} & \multirow{2}{*}{GiraffeZebraID} & MiewID & \multirow{2}{*}{0.1} & 82.1 & 82.9 & 31.7 & 27.2 & 38.6 & 82.0 \\
 & & MegaD & & 92.4 & 85.1 & 31.4 & 16.3 & 33.9 & 85.1 \\
\midrule
\multirow{2}{*}{19} & \multirow{2}{*}{Giraffes} & MiewID & \multirow{2}{*}{0.6} & 98.3 & 97.4 & 39.9 & 30.2 & 40.3 & 97.4 \\
 & & MegaD & & 98.4 & 97.2 & 42.1 & 20.5 & 42.4 & 97.2 \\
\midrule
\multirow{2}{*}{20} & \multirow{2}{*}{HappyWhale} & MiewID & \multirow{2}{*}{0.1} & 45.2 & 35.3 & 32.4 & 23.4 & 71.7 & 35.3 \\
 & & MegaD & & 63.1 & 31.8 & 29.7 & 15.5 & 47.0 & 31.7 \\
\midrule
\multirow{2}{*}{21} & \multirow{2}{*}{HumpbackWhaleID} & MiewID & \multirow{2}{*}{0.1} & 88.8 & 83.6 & 72.1 & 70.5 & 81.2 & 83.6 \\
 & & MegaD & & 88.2 & 84.8 & 28.4 & 39.7 & 32.2 & 84.8 \\
\midrule
\multirow{2}{*}{22} & \multirow{2}{*}{IPanda50} & MiewID & \multirow{2}{*}{2.5} & 56.1 & 46.3 & 52.3 & 19.1 & 92.9 & 44.9 \\
 & & MegaD & & 97.3 & 78.8 & 67.3 & 17.2 & 68.3 & 78.3 \\
\midrule
\multirow{2}{*}{23} & \multirow{2}{*}{LionData} & MiewID & \multirow{2}{*}{1.0} & 10.2 & 9.1 & 3.1 & 1.6 & 23.2 & 8.2 \\
 & & MegaD & & 43.1 & 18.2 & 7.5 & 1.6 & 15.5 & 17.4 \\
\midrule
\multirow{2}{*}{24} & \multirow{2}{*}{MacaqueFaces} & MiewID & \multirow{2}{*}{2.9} & 98.9 & 98.2 & 15.3 & 12.3 & 12.8 & 98.2 \\
 & & MegaD & & 99.4 & 99.0 & 26.2 & 12.0 & 24.1 & 99.0 \\
\midrule
\multirow{2}{*}{25} & \multirow{2}{*}{NDD20} & MiewID & \multirow{2}{*}{1.6} & 45.8 & 29.9 & 35.4 & 19.1 & 76.4 & 28.7 \\
 & & MegaD & & 85.2 & 35.6 & 61.4 & 17.2 & 71.5 & 34.5 \\
\midrule
\multirow{2}{*}{26} & \multirow{2}{*}{NOAARightWhale} & MiewID & \multirow{2}{*}{0.3} & 28.9 & 9.6 & 20.5 & 4.5 & 70.6 & 9.4 \\
 & & MegaD & & 53.2 & 13.2 & 37.5 & 4.5 & 70.3 & 12.9 \\
\midrule
\multirow{2}{*}{27} & \multirow{2}{*}{OpenCows2020} & MiewID & \multirow{2}{*}{3.7} & 99.1 & 98.4 & 73.3 & 65.2 & 73.0 & 98.4 \\
 & & MegaD & & 100.0 & 99.0 & 70.9 & 52.6 & 69.9 & 99.0 \\
\midrule
\multirow{2}{*}{28} & \multirow{2}{*}{SMALST} & MiewID & \multirow{2}{*}{10.0} & 98.1 & 97.2 & 46.3 & 51.5 & 47.1 & 96.9 \\
 & & MegaD & & 100.0 & 99.7 & 39.5 & 35.3 & 32.8 & 99.7 \\
\midrule
\multirow{2}{*}{29} & \multirow{2}{*}{ZindiTurtleRecall} & MiewID & \multirow{2}{*}{0.1} & 42.9 & 43.6 & 6.6 & 3.3 & 15.2 & 42.9 \\
 & & MegaD & & 88.6 & 84.9 & 7.5 & 2.6 & 8.4 & 84.9 \\
\midrule
\multirow{2}{*}{30} & \multirow{2}{*}{CzechLynx} & MiewID & \multirow{2}{*}{1.9} & 75.6 & 68.0 & 52.8 & 25.7 & 69.0 & 67.4 \\
 & & MegaD & & 61.3 & 50.2 & 50.3 & 23.0 & 81.5 & 49.2 \\
\midrule
\multirow{2}{*}{31} & \multirow{2}{*}{BrownBearHeads} & MiewID & \multirow{2}{*}{1.6} & 96.5 & 96.0 & 62.6 & 58.2 & 64.2 & 95.9 \\
 & & MegaD & & 92.1 & 92.4 & 68.4 & 43.9 & 73.8 & 92.0 \\
\midrule
\multirow{2}{*}{32} & \multirow{2}{*}{WhaleSharkID} & MiewID & \multirow{2}{*}{0.5} & 88.7 & 82.0 & 14.3 & 12.8 & 15.6 & 81.9 \\
 & & MegaD & & 83.4 & 70.9 & 17.2 & 9.7 & 20.1 & 70.7 \\
\midrule
\multirow{2}{*}{33} & \multirow{2}{*}{LeopardID2022} & MiewID & \multirow{2}{*}{1.1} & 92.5 & 89.1 & 38.9 & 24.4 & 41.3 & 89.0 \\
 & & MegaD & & 78.4 & 59.1 & 41.9 & 21.8 & 52.8 & 58.6 \\
\midrule
\multirow{2}{*}{34} & \multirow{2}{*}{Saimaa seal} & MiewID & \multirow{2}{*}{0.7} & 43.9 & 33.0 & 31.3 & 11.5 & 70.9 & 32.5 \\
 & & MegaD & & 41.5 & 33.5 & 32.6 & 9.6 & 78.3 & 33.0 \\
\bottomrule
\end{tabular}
}
\captionof{table}{\textbf{CFR evaluation for all datasets.}
Values are percentages. The \# column matches Fig.~\ref{fig:dataset_evidence};
Full, Frg., SC, and Sil are retrieval on the full-image, foreground-only,
shortcut-only, and silhouette-only views, respectively. SSI and CFR follow
Sec.~\ref{sec:cfr-score}.}
\label{tab:supp-all-cfr-components}
\end{minipage}

\end{document}